\documentclass[conference]{IEEEtran}
\IEEEoverridecommandlockouts
\usepackage{hyperref}
\usepackage{cite}
\usepackage{subfig}
\usepackage{booktabs} 
\usepackage{amsmath,amssymb,amsfonts}
\usepackage{algorithmic}
\usepackage{graphicx}
\usepackage{textcomp}
\usepackage[table]{xcolor}

\newcommand{\M}{{Fair-Gender}}
\def\BibTeX{{\rm B\kern-.05em{\sc i\kern-.025em b}\kern-.08em
    T\kern-.1667em\lower.7ex\hbox{E}\kern-.125emX}}
\begin{document}

\title{
Bridging the Fairness Gap: Enhancing Pre-trained Models with LLM-Generated Sentences
\thanks{*Corresponding author.}
}

\author{\IEEEauthorblockN{Liu Yu, Ludie Guo, Ping Kuang*, Fan Zhou}
\IEEEauthorblockA{\textit{University of Electronic Science and Technology of China, Chengdu, Sichuan 610054, China} \\
liu.yu@std.uestc.edu.cn; 202222090414@std.uestc.edu.cn; kuangping@uestc.edu.cn; fan.zhou@uestc.edu.cn
}
}

\maketitle

\begin{abstract}
Pre-trained language models (PLMs) are trained on data that inherently contains gender biases, leading to undesirable impacts. Traditional debiasing methods often rely on external corpora, which may lack quality, diversity, or demographic balance, affecting the effectiveness of debiasing. With the rise of large language models and their extensive knowledge, we propose enhancing fairness (\M) in PLMs by absorbing coherent, attribute-balanced, and semantically rich sentences. However, these sentences cannot be directly used for debiasing due to alignment issues and the risk of negative transfer. We address this by applying causal analysis to estimate causal effects, filtering out unaligned sentences, and identifying aligned ones for incorporation into PLMs, thereby ensuring positive transfer. Experiments show that our approach significantly reduces gender biases in PLMs while preserving their language expressiveness.
\end{abstract}


\section{Introduction}
Lightweight pre-trained language models such as BERT~\cite{bert} and RoBERTa~\cite{roberta} have achieved remarkable advancements across a wide range of tasks~\cite{sun2019bias,elahi2021investigating,shen2023towards}, including language understanding~\cite{meng2022generating}, document classification~\cite{bhardwaj2021investigating}, and multitask text generation. Their effectiveness largely stems from the ability of these PLMs to generate contextual representations. However, due to out-of-distribution~\cite{lu2022rationale,veselovsky2305generating} or stereotypical biases inherent in the training corpus, their extensive deployment may inadvertently perpetuate biased or stereotypical representations~\cite{WEAT,kolisko2023exploring}, leading to potential unfairness in applications~\cite{mok2023people} involving diverse social demographic groups. For instance, gender bias is evident when PLMs are more likely to associate \textit{male} (\textit{female}) attributes with \textit{programmers} (\textit{nurses}). This issue is particularly critical in specialized fields like law, medicine, or human resources~\cite{jatoba2019evolution}, where ensuring fairness in encoded representations becomes crucial.


\textbf{Related works and limitations.} Many solutions~\cite{yu2023mixup,mao2023debiasing,yu2024biases} have been proposed to mitigate social biases in PLMs. Based on whether the contextualized debiasing methods are directly integrated with downstream tasks, external corpora-based methods can be categorized into two types: (1) \textit{Task-Agnostic} methods: Sent-Debias~\cite{sent-debias} and FairFil~\cite{fairfil} are \textit{post-hoc} methods that keep PLM parameters unchanged. ADEPT~\cite{yang2023adept} introduces a novel training criterion that optimizes only the continuous prompt parameters while keeping the base model frozen. Auto-Debias~\cite{auto-debias}, Context-Debias~\cite{context-debias}, and MABEL~\cite{he2022mabel} eliminate biases in PLMs through \textit{fine-tuning} with various bias-neutralizing loss functions. (2) \textit{Task-Aware} methods focus on preventing bias from re-emerging when applying debiased models in real-world applications. For instance, Causal-Debias~\cite{causal-debias} integrates debiasing with downstream fine-tuning via causal invariant learning, while Gender-tuning~\cite{gender-tuning} provides a debiasing mechanism for any PLM, using standard fine-tuning techniques. Despite their notable success, all these methods rely on external corpora to identify and mitigate biases, aiming to ensure adequate demographic diversity. 
Moreover, acquiring high-quality corpora is typically expensive, and noisy information may be introduced~\cite{zheng2023preserving}, leading to insufficient bias reduction. 

\textbf{Motivation.} Recent foundational models like ChatGPT~\cite{chiang2023vicuna} and LLaMa~\cite{touvron2023llama} have demonstrated impressive intelligence across a range of complex tasks, including vision-based question answering~\cite{liang2024toa} and knowledge reasoning~\cite{li2024enhanced}, among others. Given that language models can incorporate extensive knowledge during pre-training~\cite{bian2023chatgpt}, we consider LLMs as valuable knowledge bases that can provide insights to enhance fairness in other lightweight PLMs. The limitations in the quality, diversity, and balance of external corpora, combined with the notable knowledge capabilities of LLMs~\cite{bian2023chatgpt}, motivate us to leverage coherent, attribute-balanced, and semantically rich sentences from LLMs to debias lightweight PLMs.


\textbf{Present work.} In our approach, we limit the prompts for LLMs to ensure that the generated sentences are focused on social aspects. While these sentences are semantically rich and balanced across various groups, not all are immediately suitable for debiasing due to the significant differences in the latent spaces of LLMs and PLMs, which arise from their distinct pre-training data. Consequently, the sentences extracted from LLMs might include content that is difficult for PLMs to comprehend, and using it indiscriminately for debiasing could result in negative transfer~\cite{zheng2023preserving}. To address this, we adopt a causal perspective and estimate causal effects to identify sentences that are well-aligned for debiasing PLMs. Sentences with strong causal effects are considered aligned sentences, facilitating positive transfer and reducing the learning burden on PLMs by filtering out unaligned sentences, thereby preventing model degradation.

\section{Preliminary}
\textbf{Problem Definition.}
Let $\mathcal{W}_{a}=\{(a_{1}^{(1)}, a_{2}^{(1)}, \cdots, a_{d}^{(1)}), \\(a_{1}^{(2)}, a_{2}^{(2)}, \cdots, a_{d}^{(2)}), \cdots\}$ denote \textit{attribute} words composed of multiple $d$-tuple and $\mathcal{W}_{t}=\{v_1, v_2, \cdots\}$ denotes \textit{target} words, respectively. In the case of binary gender ($d=2$), 
attribute words are gender-specific pairs:
(\textit{airmen}, \textit{airwomen}), (\textit{princes}, \textit{princesses}),  (\textit{king}, \textit{queen}),
the target words are gender-neutral (e.g., \textit{captain, boss, professor}). 
We denote a pre-trained language model as $\mathcal{M}$, and our goal is to fine-tune $\mathcal{M}$ via the generated pairwise sentences (containing gender-specific pairs and target words) to mitigate unwanted stereotypical associations to obtain the fine-tuned model $\mathcal{M}'$.


\section{Methodology}
\label{Methodology}
\begin{figure}[th]
    \centering
    \includegraphics[width=0.95\columnwidth]{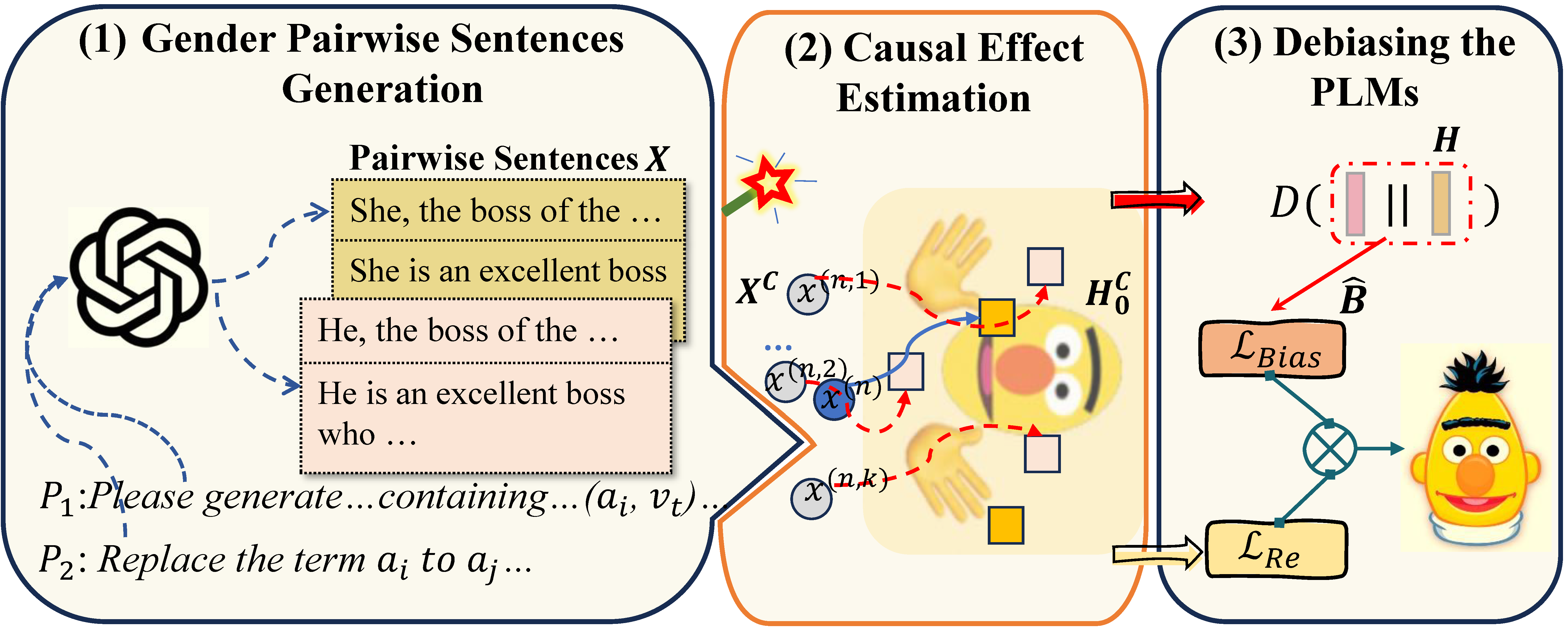}
    \caption{Our comprehensive debiasing framework \M. 
    }
    \label{fig:overall-frame}
\end{figure}

In this section, we address the process of generating knowledge using large-scale model and explore strategies for optimizing the utilization of the generated sentences. The comprehensive framework is illustrated in Fig.~\ref{fig:overall-frame}.

\textbf{Gender Pairwise Sentences Generation from LLMs.} For prompting LLM, we use two well-designed system prompts $P_1$, $P_2$
(depicted in step 1 in Fig.~\ref{fig:overall-frame}) 
to automatically generate pairwise sentences in two steps:
\begin{itemize} 
    \item $P_1$: \textit{Please generate ten sentences containing the words in a tuple ($a_{i}$, $v_{t}$) simultaneously. Control the word count in every generated sentence to around 20. The generated sentences strive for creativity, diversity, and logic.}
    \item $P_2$: \textit{Replace the term $a_{i}$ to $a_{j}$, and correct personal pronouns in the above generated ten sentences.}
\end{itemize}
wherein $a_i\in \{a_{1}, a_{2}, \cdots, a_{d}\}$, 
and $a_j$ used for replacement comes from other $d-1$ elements,
and $v_t \in \mathcal{W}_t$.
The generated sentences, containing $(a_i, \underline{v_t})$, $(a_j, \underline{v_t})$, are with same target words $\underline{v_t}$ over $d$-tuple attribute words, for instance (\textit{she}, \textit{\underline{boss}}), (\textit{he} and \textit{\underline{boss}}).
We can generate extensive pairwise sentences with high-quality, diversity, and gender balance in this step.

\begin{figure}[h]
    \centering
    \includegraphics[width=0.85\columnwidth]{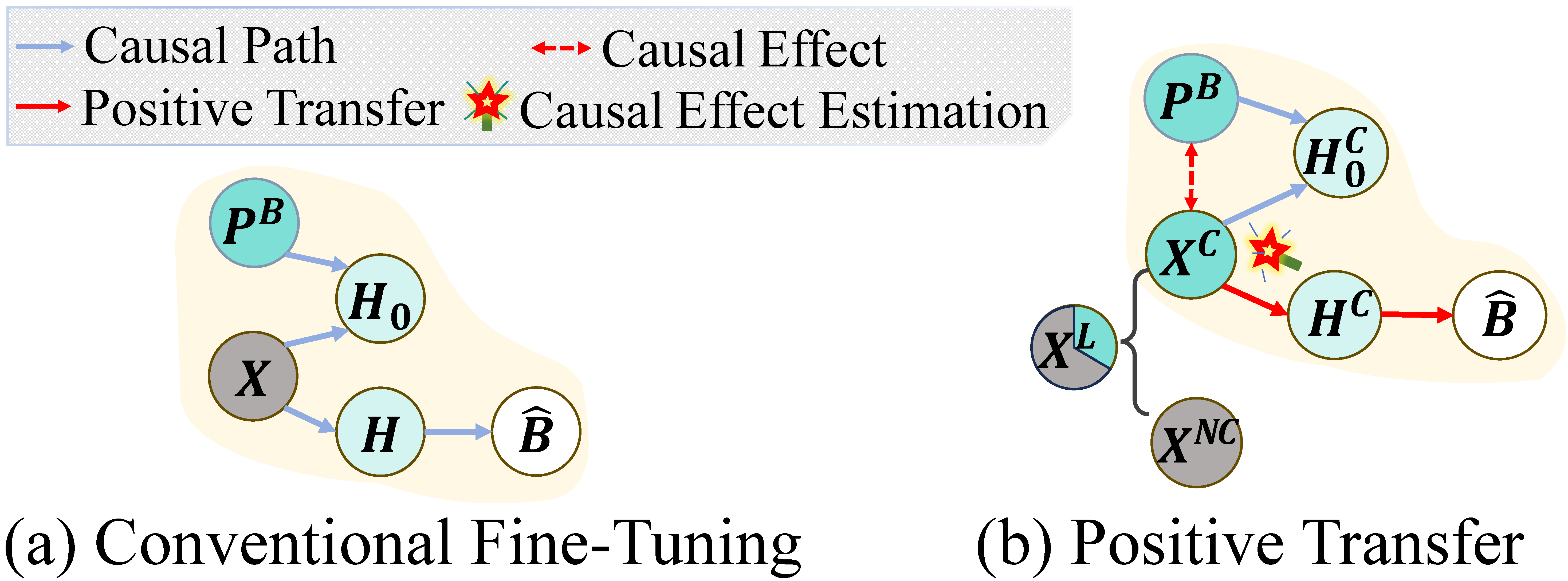}
    \caption{The comparison of structural causal model between conventional methods and our \M. 
    }
    \label{fig:scm}
\end{figure}

\textbf{Debiasing PLMs via Generated Pairwise Sentences.}
We construct a Structural Causal Model (SCM) to capture the causality between data, models, and hidden variables for positive transfer. As shown in Fig.~\ref{fig:scm}, the biased pre-trained data of PLMs is denoted as $P^B$, pairwise sentences as $X$, hidden variables from the initial and fine-tuned models as $H_0$ and $H$, and the predicted bias magnitude as $\hat{B}$. The causal associations are: (1) $X \rightarrow H \rightarrow \hat{B}$: where $H$ is derived from $X$ by PLMs, and $\hat{B}$ is measured from $H$; (2) $X \rightarrow H_0 \leftarrow P$: where $H_0$ is influenced by both $P$ and $X$.

As shown in Fig.~\ref{fig:scm} (a), due to the significant difference between LLMs and PLMs in pre-training data creating disparities in their latent spaces, applying LLM-derived knowledge directly for bias mitigation can cause negative transfer~\cite{zheng2023preserving}. 

To achieve a positive transfer of generated sentences $X^L$ from LLM and enhance the debiasing impact of PLMs, it is crucial to establish an alignment between the hidden space of PLMs and LLM.    
As shown in Fig.~\ref{fig:scm} (b), we split $X^L$ into two nodes $X^{C}$ and $X^{NC}$.
$X^C$ represents the samples where causal effects are significant, and they should align with PLMs to enhance their fairness. 
$X^{NC}$ signifies samples with smaller causal effects, presenting unaligned sentences due to negative transfer -- we filter this part of the data.
In summary, the fine-tuned PLMs absorb LLM's demographic Sentences by utilizing causal effects ($P^B \leftrightarrow X^{C}$ ). When conditioning on $H^C_0$, the final bias magnitude depends on the degree of assimilating aligned knowledge from causal paths $P^B \leftrightarrow X^C \rightarrow H^C \rightarrow  \hat{B}$ (positive transfer).

\textbf{Causal Effect Estimation.}
When predicting $\hat{B}^{(n)}$, we first obtain the initial hidden state $h^{(n)}_0 = \mathcal{M}_0(x^{(n)})$. $H_0 = h^{(n)}_0$ means $X$ represents all samples with hidden features $h^{(n)}_0$. However, due to high-dimensional sparsity, the only suitable candidate is $x^{(n)}$. 
Following~\cite{zheng2023preserving}, we relax this constraint and use the joint estimation of K-Nearest-Neighbor (KNN) samples to estimate causal effect $\Delta$ between $P$ and $X^C$:
\begin{align}\label{collid_eff}
 & \Delta=  \sum_{n=1}^N \delta^{(n)} \approx \\
 &   \sum_{n=1}^N \sum_{k=0}^{k_n}  B\left(\mathcal{M}_{H}(X^C=x^{(n, k)})\right) {S}_H\left(x^{(n)}, x^{(n, k)}\right)\nonumber
\end{align}
where $N$ is the total number of pairwise sentences from aligned knowledge $X^C$, and $k_n$ is the number of KNNs for the $n$-th pairwise sentence in estimating $\hat{B}^{(n)}$. $x^{(n, k)}$ is the $k$-th nearest neighbor of $x^{(n)} \in \mathcal{X}^{N}$.
${S}_H\left(\cdot, \cdot \right)$ is the similarity function between $x^{(n)}$ and $x^{(n, k)}$ (denoted as ${S}_{n,k}$), with $\sum_{k=0}^{k_n} {S}_H\left(x^{(n)}, x^{(n, k)}\right)=1$. Since $x^{(n)}$ is most similar to itself, $x^{(n, 0)}$ is set as the anchor sample when $k = 0$. $B\left(\mathcal{M}_{H}(X^C=x^{(n, k)})\right)$ is the bias prediction of $\hat{B}^{(n)}$ when $x^{(n,k)}$ is the input to model $\mathcal{M}_H$. Equation~\eqref{collid_eff} shows that the total causal effect $\Delta$ is the sum of $N$ aligned sentences' causal effects $\delta^{(n)}$, with each $\delta^{(n)}$ approximated by the weighted sum of the bias prediction when the input is the anchor sample $x^{(n)}$ and its KNNs. 
The debiasing objective is:
\begin{align}\label{equ:bias_loss}\nonumber
\mathcal{L}_{b}= \sum_{ x^{(n)}\in \mathcal{X}^C}\sum_{k=0}^{k_n}\mathcal{D}\big(X^C=x^{(n,k)}\big)S_{n,k}
\end{align} 
where ${\mathcal{L}_{b}}$ is a rewrite of the causal effect $\Psi$ estimated from $X^C$. To integrate the aligned knowledge $X^C$ into PLMs, we assess the knowledge preservation strength for each pairwise sentence $x^{(n)}$ by selecting its KNNs $x^{(n,k)}$. $\mathcal{D}_1$ measures the relative JSD between sentences with attribute words and those with target words, defined as:

\begin{align}\nonumber
\mathcal{D}\left(x^{(n)}\right)=\sum_{i, j \in\{1, \ldots, d\}, i<j}\left\{J SD\left(R^{x_i^{(n)}} \| R^{x_j^{(n)}}\right)\right\}
\end{align}
where $R^{x_i^{(n)}}$/$R^{x_j^{(n)}}$ measures the distance from sentence $x_i^{(n)}$ with attribute words $a(i)$/$a(j)$ to sentences with all target words, respectively. 
The optimization of $\mathcal{L}_{b}$ ensures that the pairwise attribute words $a(i)$ and $a(j)$ have a uniform distance to all neutral target words, satisfying the fairness criterion.




To prevent harm to PLM expressiveness from full parameter fine-tuning, we add an auxiliary representation loss $\mathcal{L}_{r}$ to preserve its inherent language modeling capability, defined as:

\begin{equation}\label{equ:rep loss}
    \mathcal{L}_{r}= MSE(\mathcal{M}_{H}(\cdot)||\mathcal{M}^{\prime}_{H}(\cdot))
\end{equation}
where $\mathcal{L}_{r}$ uses Mean Squared Error to measure the difference between the original model’s hidden states $\mathcal{M}{H}$ and the debiased model’s $\mathcal{M}_{H}^{\prime}$, aiming to minimally adjust the PLM’s parameters. The overall training loss is minimized as follows:
\begin{align*}
    \mathcal{L} =  \mathcal{L}_{b}  + \lambda \cdot \mathcal{L}_{r},
\end{align*}\label{equ:overall_loss} 
wherein $\mathcal{L}_{r}$ is tempered by the hyper-parameter $\lambda$.

\section{Experiments}
\noindent \textbf{Comparison.} 
Our benchmarks 
including \textit{Task-Agnostic} models: Context-Debias~\cite{context-debias}, Auto-Debias~\cite{auto-debias}, FairFil~\cite{fairfil}, and MABEL~\cite{he2022mabel}; and \textit{Task-Aware} models: Gender-tuning~\cite{gender-tuning} and Causal-Debias~\cite{causal-debias}. We utilize GPT-3.5-turbo API as the source LLM for pairwise sentences generation. Three PLMs are as backbones: BERT~\cite{bert}, ALBERT~\cite{albert}, and RoBERTa~\cite{roberta}. 
Following prior studies, we choose the gender/racial/religion word lists from \cite{context-debias}, \cite{manzini2019black}, and \cite{sent-debias} respectively.

\noindent \textbf{Evaluation.}  We report bias indicators including SEAT (the absolute value closer to 0 means lower biases)~\cite{SEAT}, StereoSet \cite{nadeem2020stereoset} (LM, SS, ICAT), and CrowS-Pairs \cite{crowspair}. 
We evaluate our framework on three GLUE tasks to measure the model's expressiveness, including SST-2, CoLA, and QNLI. 

\begin{table}[ht]
\small
\centering
\resizebox{0.48\textwidth}{!}{
\begin{tabular}{l|c|ccc|c}

\toprule
\multicolumn{2}{c|}{} & \multicolumn{3}{c|}{\textbf{StereoSet}} & \multicolumn{1}{c}{}  \\
\cmidrule{3-5}

\multicolumn{1}{c}{\textbf{Methods}} & \multicolumn{1}{c|}{\textbf{SEAT}}  & LM$\uparrow$ & SS $\diamond$ & \multicolumn{1}{c|}{ICAT $\uparrow$} &  \multicolumn{1}{c}{\textbf{CrowS-P*$\diamond$}} \\

\cmidrule{1-6}
\rowcolor{gray!20}\textbf{BERT} & 0.35
& 84.17  & 60.28
& 66.86  & 57.25 \\
\textsc{+Context-D*} & 0.53 
& 85.42  & 59.35 
& 69.45  & 58.01 \\
\textsc{+FairFil} 
& 0.15 
& 44.85  & \textbf{50.93} 
& 44.01  & 49.07 \\
\textsc{+Auto-D*} & 0.14 
& 74.08  & 52.88
& 69.81  & 54.92 \\
\textsc{+MABEL} & 0.582 
& 84.80  & 56.92 
& 73.07  & 50.76 \\

%
+Ours  & \textbf{0.068}  
& \textbf{85.30} & 55.34 
& \textbf{74.91} & \textbf{50.43} \\
\cmidrule{1-6}
\rowcolor{gray!20}\textbf{ALBERT} & 0.28
& 90.73  & 63.58
& 66.09  & 56.87 \\
\textsc{+Context-D*} & 0.33 
& 91.02  & 60.23
& 72.40  & 53.91 \\
\textsc{+Auto-D*} & 0.18 
& 88.43  & 61.76
& 67.62  & 47.86 \\
+Ours & \textbf{0.16} 
& \textbf{91.30}  & \textbf{59.14} 
& \textbf{74.63} & \textbf{47.97} \\
\cmidrule{1-6}
\rowcolor{gray!20}\textbf{RoBERT} & 0.67
& 71.75  & 53.65
& 66.50  & 54.96  \\
\textsc{+Context-D*} & 1.09 
& 70.85  & 54.74
& 64.13  & 59.48\\
\textsc{+Auto-D*} & 0.20 
& 69.85  & 54.21
& 63.13 & 49.77 \\
+Ours & \textbf{0.15} 
& \textbf{72.23}  & \textbf{53.54} 
& \textbf{67.16}  & \textbf{50.45} \\

\bottomrule
\end{tabular}
}
\caption{Gender debiasing results. $\textsc{*}$: abbrev for a model or metric. $\uparrow$: the larger, the better. $\diamond$: the closer to 50, the better.
}
\label{tab:main-table}
\end{table}

\noindent \textbf{Configuration.}
We train models in 4 epochs with learning rate $5\times e^{-5}$ on
a single GeForce RTX 3090 GPU.

\begin{figure}[t]
    \centering
    \begin{minipage}{0.5\linewidth}
        \centering
        \subfloat[Toxicity detection.]{
            \includegraphics[height=0.62\linewidth]{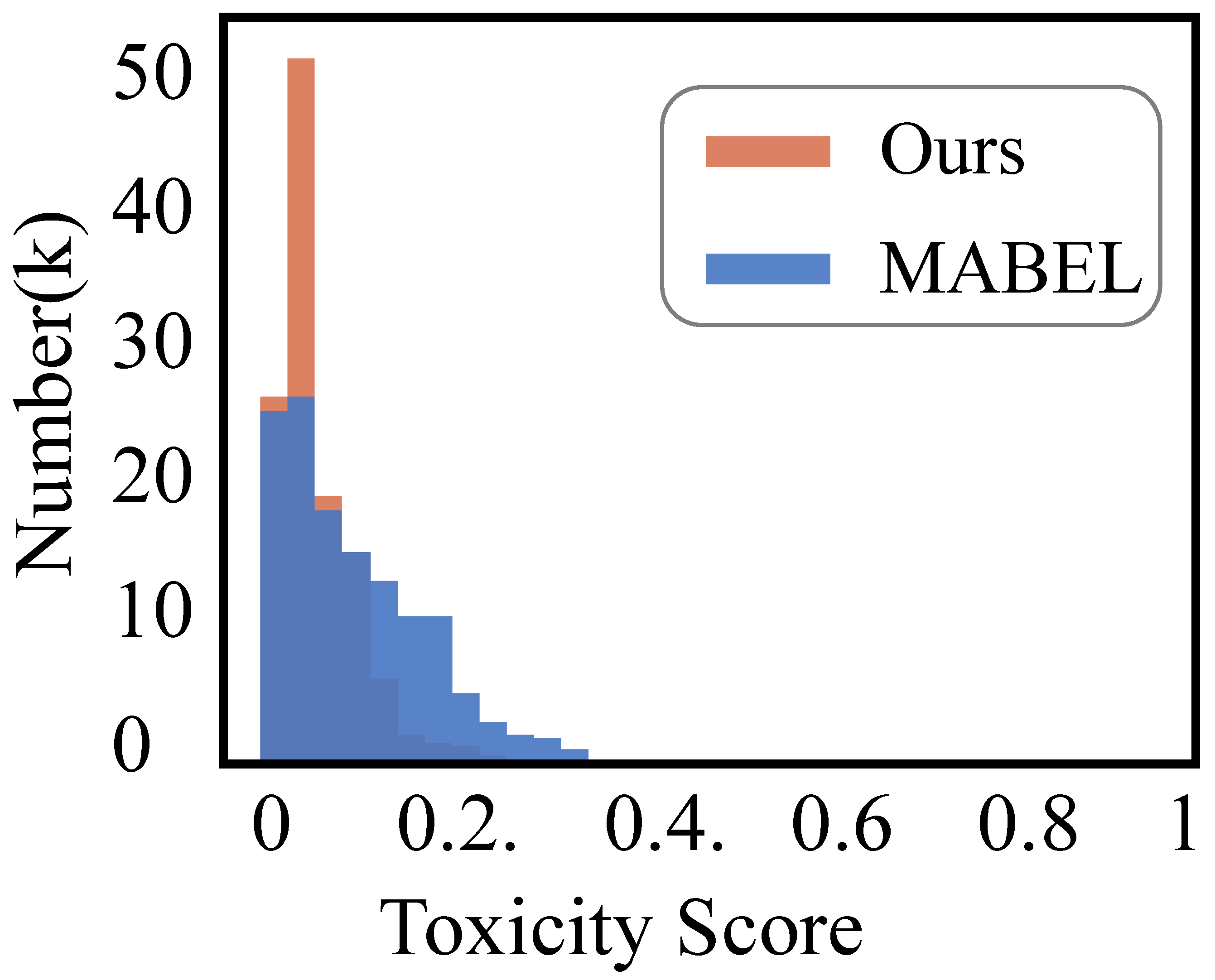}
            \label{fig:ablation-a}
        }
    \end{minipage}
     \hfill
    \begin{minipage}{0.48\linewidth}
        \centering
        \subfloat[Results of three ablation versions.]{
            \includegraphics[height=0.68\linewidth]{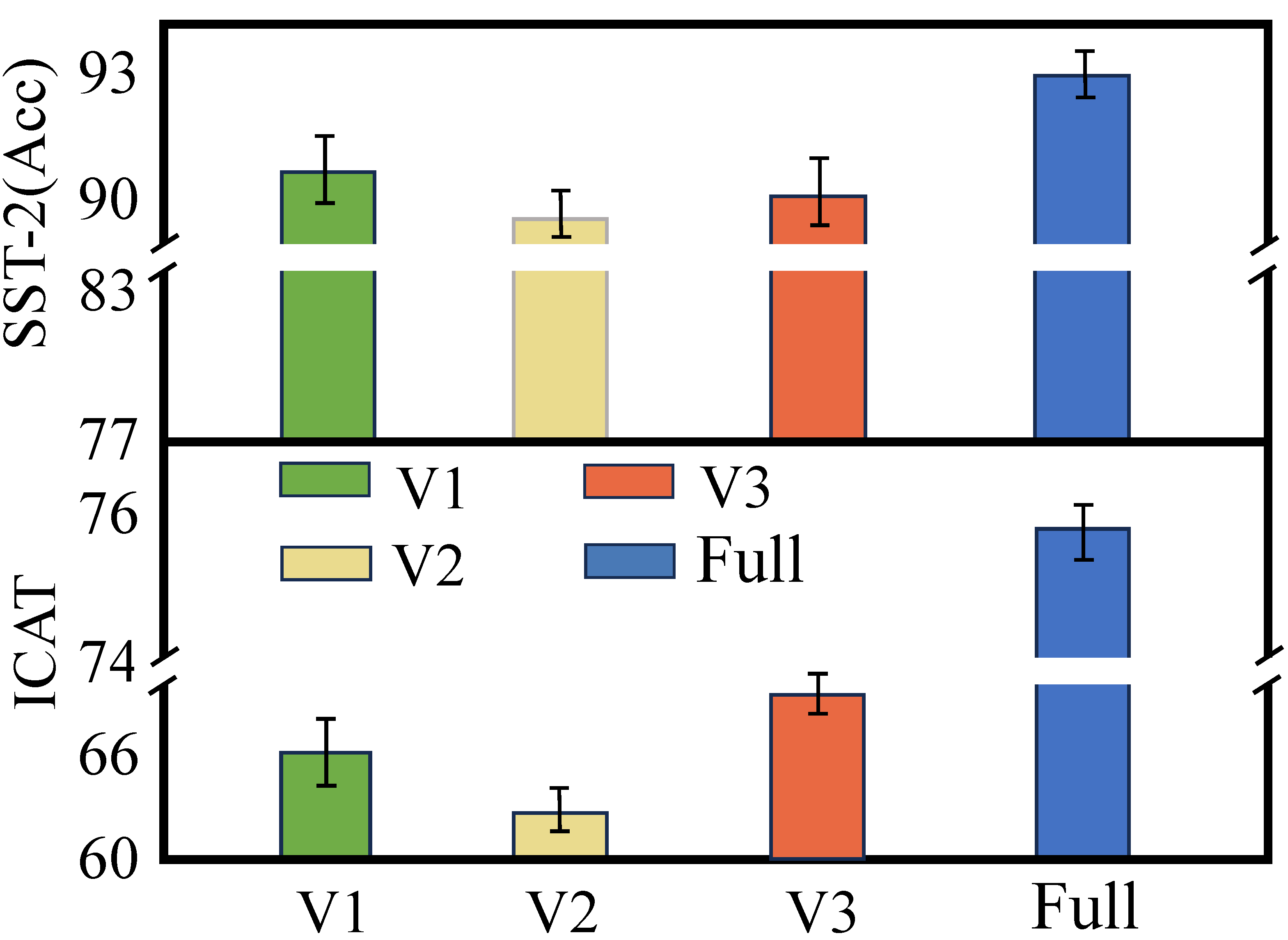}
            \label{fig:ablation-b}
        }
    \end{minipage}
    \caption{Toxicity detection of sentences, and ablation versions.}
    \vspace{-1em}
    \label{fig:ablation}
\end{figure}

\noindent \textbf{Toxicity Detection.} 
Ensuring the harmlessness of generated sentences is essential for the debiasing process.
We apply the Comprehend API from Amazon Web Services (AWS) for toxicity detection. 
The toxicity scores in the $x$-axis range from $0$ to $1$, where a higher score indicates a greater likelihood of the text containing toxic content.
To have a fair comparison, we choose MABEL's entailment data in the gender domain. The toxicity distribution in Fig.~(\ref{fig:ablation-a}) is reported using equivalent data volume.
The toxicity distribution shows that the low-toxicity segment (the peak in blue closer to $0$) of \M~is notably lower than MABEL, and \M~exhibits significantly fewer high-toxicity levels (closer to $1$).
Given the potential constraints of the toxicity detection tool and aiming to improve the data quality for debiasing, we select 60\% of these samples with the lowest toxicity across three domains for experiments, instead of directly using all the data as MABEL. 

\noindent \textbf{Overall Performance.} As indicated by the remarkable ICAT metric score in Table~\ref{tab:main-table}, our \M~strikes a favorable balance between language expressiveness and fairness. Notably, \M~even exhibits a slight improvement in LM metrics compared to the backbones, e.g., with the score rising from $84.17$ to $85.30$ in BERT.
For the SEAT value, \M~achieves the best score, and improves $0.072$ compared to the SOTA model Auto-Debias. 
Additionally, \M~outperforms others in CrowS-Pairs with the best score of $50.43$.
\M~performs best in ALBERT and RoBERT,
while it does not rank top in terms of the SS value on BERT model, we note that this metric should be considered alongside LM, rather than evaluated in isolation.
For instance, FairFil achieves the highest SS, yet its language modeling capability, as indicated by the lower LM score, suffers a marked decline and trails other methods.

\noindent \textbf{Ablation Study.}\label{ablation}
To verify the effectiveness of \M, we consider the following ablated version: 
\begin{itemize}
    \item 
    (V1) \texttt{w/o} $\mathcal{L}_{r}$: Remove the designed representation preserving loss $\mathcal{L}_{r}$;
    \item (V2) Rand-$1$: Replace KNNs for causal effect estimation with randomly selected pairwise sentences; 
    \item (V3) Rand-$2$: Reduce the number of KNNs from $5$ to $2$.
    
\end{itemize}

\begin{figure}[tb]
\centering
\subfloat[Debiasing effect.]{
\label{fig:curve:a}
    \setlength{\fboxrule}{1pt}
    \setlength{\fboxsep}{0pt}
	\includegraphics[width=0.23\textwidth]{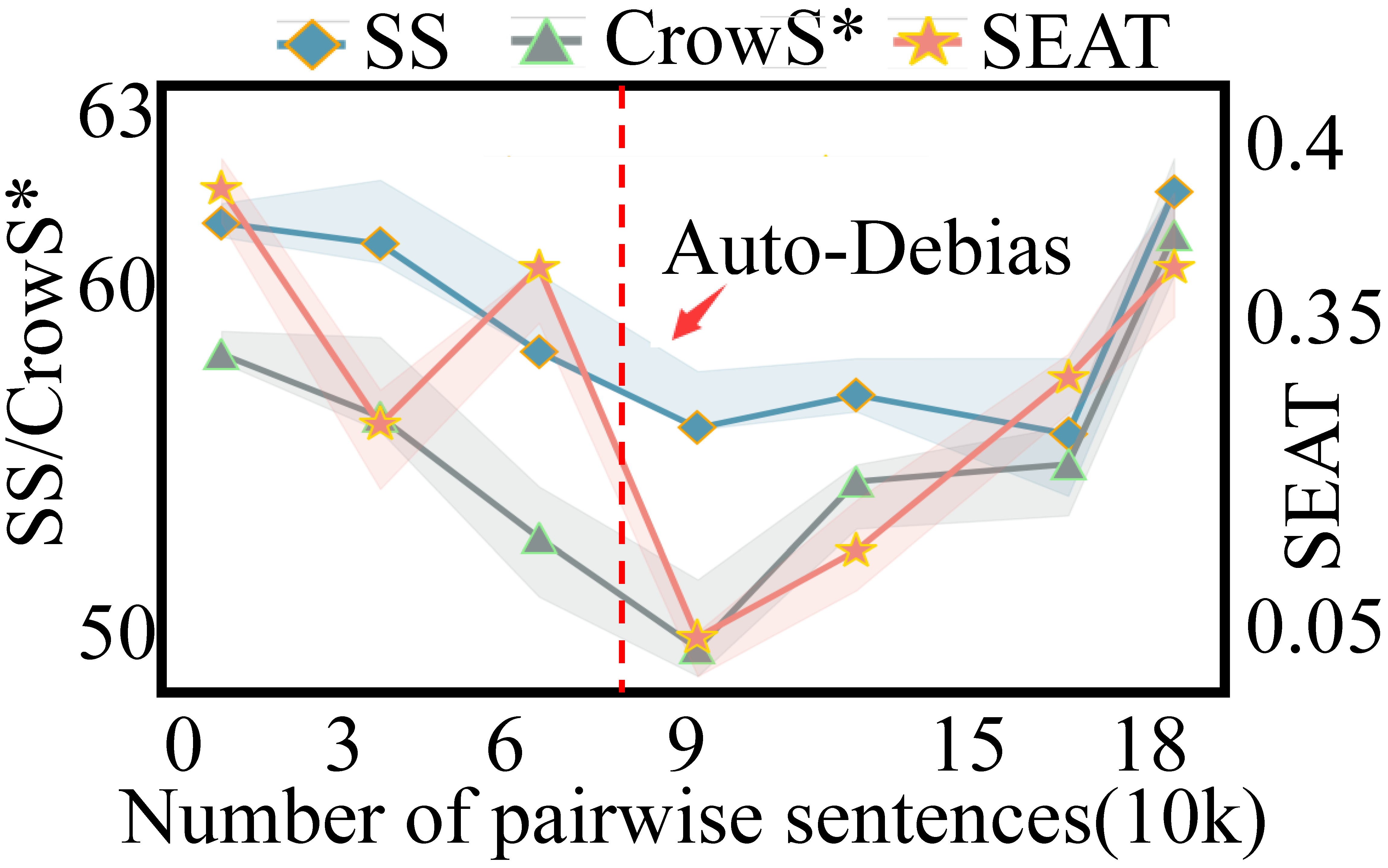}
    }
\subfloat[Model expressiveness.]{
\label{fig:curve:b}
    \setlength{\fboxrule}{1pt}
    \setlength{\fboxsep}{0pt}
	\includegraphics[width=0.230\textwidth]{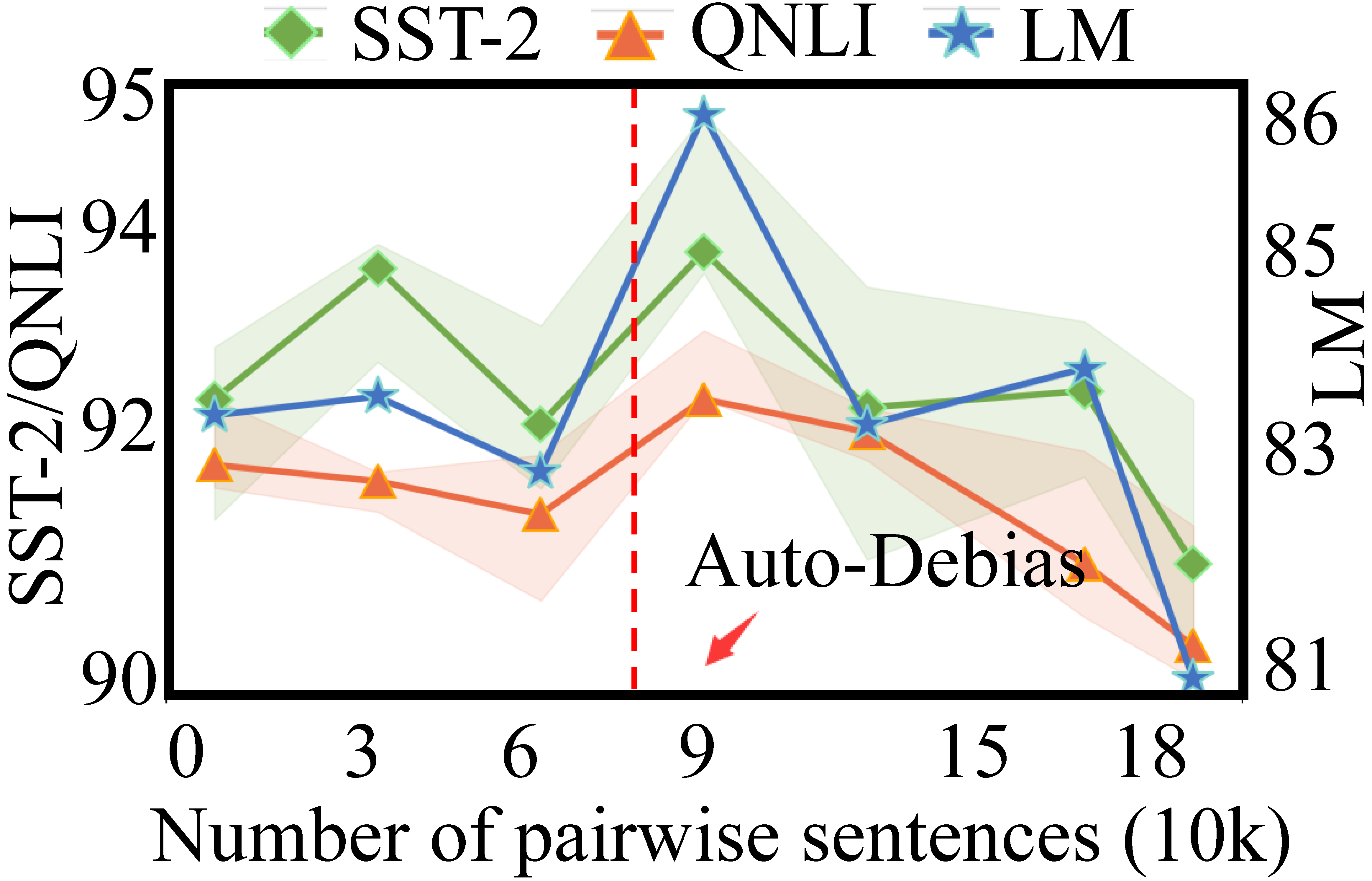}
}
\caption{The number impact of the generated pairwise sentences.
}
\label{fig:curve}
\end{figure}




Fig.~(\ref{fig:ablation-b}) shows that all variants underperform compared to the full \M~model. Removing $\mathcal{L}_{r}$ notably reduces SST-2 accuracy and ICAT value, highlighting its importance in maintaining language modeling. The lower ICAT scores in Rand-$1$ variants further emphasize the importance of KNNs in estimating the causal effect and transferring aligned knowledge regarding demographic diversity from the LLM to the PLM.
Moreover, reducing the number of KNNs still weakens the debiasing ability, indicating the importance of KNNs for knowledge positive transfer.

Additionally, we conduct an ablation study on the number of sentences to demonstrate how performance changes. As illustrated in Fig.~\ref{fig:curve}, increasing the number of sentences improves \M's fairness and language expressiveness, as shown by the debiasing effect (SS, CrowS-Pairs, SEAT metrics) in Fig.~(\ref{fig:curve:a}) and model understanding ability (LM, SST-2, QNLI tasks) in Fig.~(\ref{fig:curve:b}). However, this improvement plateaus after 90k sentences, suggesting an optimal amount for debiasing.
When compared at an equivalent number of sentences (around 80k), indicated by the red vertical dashed line in Fig.~\ref{fig:curve}, \M~significantly outperforms Auto-Debias across all metrics. This performance gap may be due to the short, biased prompts used in Auto-Debias, which lack syntax and context. In contrast, the pairwise sentences extracted from LLM offer semantically rich information across various demographic groups, making them a valuable resource.

\begin{table}[t]
\centering\small
\begin{tabular}{l|c|c|c}

\toprule
\cmidrule{1-4}

\multicolumn{1}{c|}{\textbf{Methods}} & \multicolumn{1}{c|}{\textbf{SST-2}} & \textbf{CoLA} & \textbf{QNLI}  \\
\cmidrule{1-4}
\rowcolor{gray!20}\textbf{BERT} & 92.7
& 57.6 & 91.3 \\
\textsc{+Auto-Debias} & 92.1   
& 52.1 & 91.1 \\
\textsc{+Gender-Tuning} & 92.1
& 56.6 & 91.3 \\
\textsc{+Causal-Debias} & 92.9
& 58.1 & \textbf{91.6} \\
\textsc{+\M} (ours) & \textbf{93.0} 
& \textbf{60.01} & 91.3\\

\cmidrule{1-4}
\rowcolor{gray!20}\textbf{ALBERT} & 92.6
& 58.5  & 91.3 \\
\textsc{+Auto-Debias} & 94.1 
& 58.3  & 92.1 \\
\textsc{+Gender-Tuning} & 91.7
& 58.4 & 92.1 \\
\textsc{+Causal-Debias} & 92.9
& 57.1  & 91.6 \\
\textsc{+\M} (ours) &  \textbf{93.7} 
& \textbf{58.7} &  \textbf{92.5}  \\


\bottomrule
\end{tabular}
\caption{GLUE results over benchmarks.}
\label{tab:GLUE-table}
\end{table}

\begin{figure}[t]
\centering
\subfloat[Original BERT.]{
\label{fig:a}
    \setlength{\fboxrule}{1pt}
    \setlength{\fboxsep}{0pt}
	\includegraphics[width=0.13\textwidth]
 {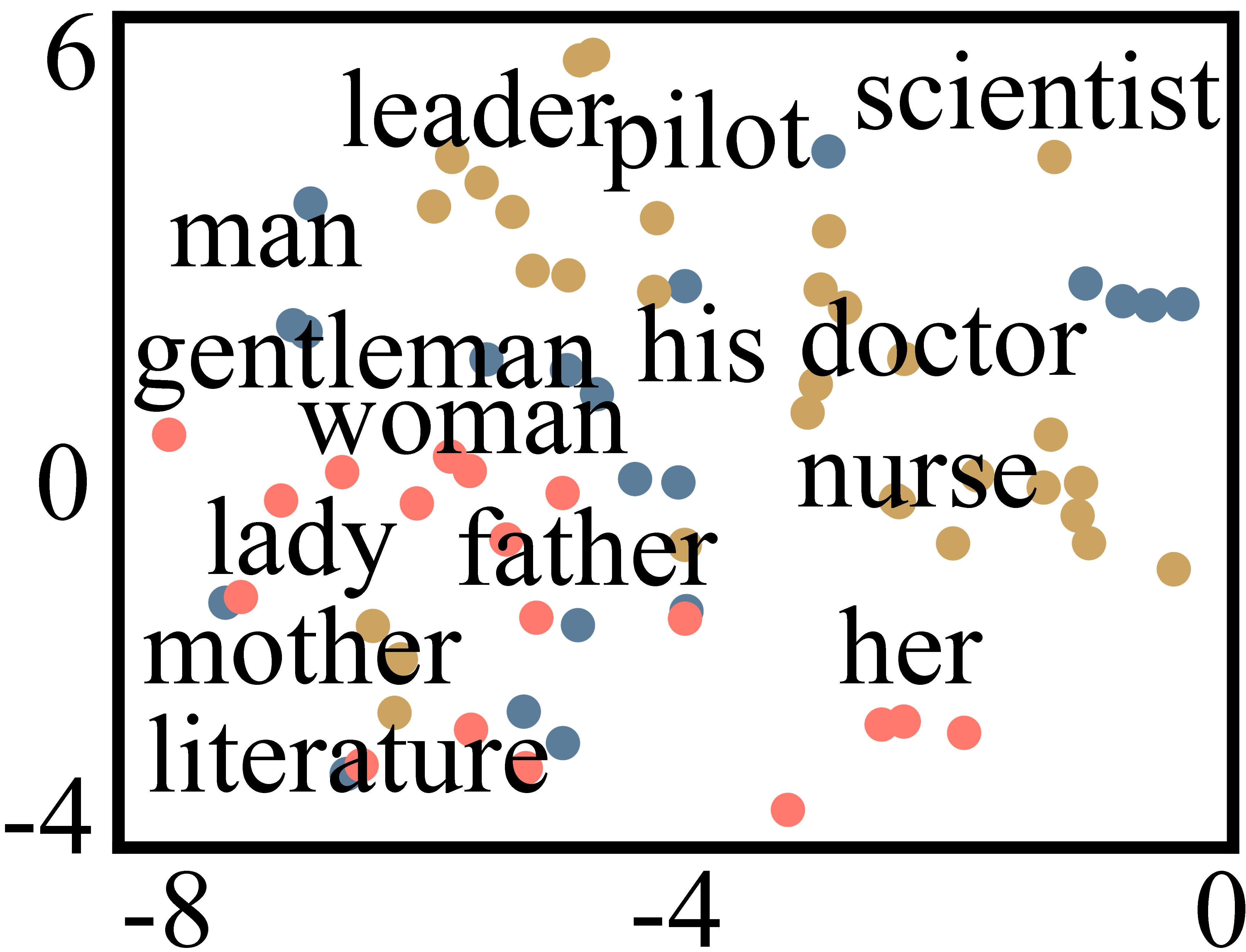}
    }
\subfloat[Auto-Debias.]{
\label{fig:b}
    \setlength{\fboxrule}{1pt}
    \setlength{\fboxsep}{0pt}
	\includegraphics[width=0.13\textwidth]{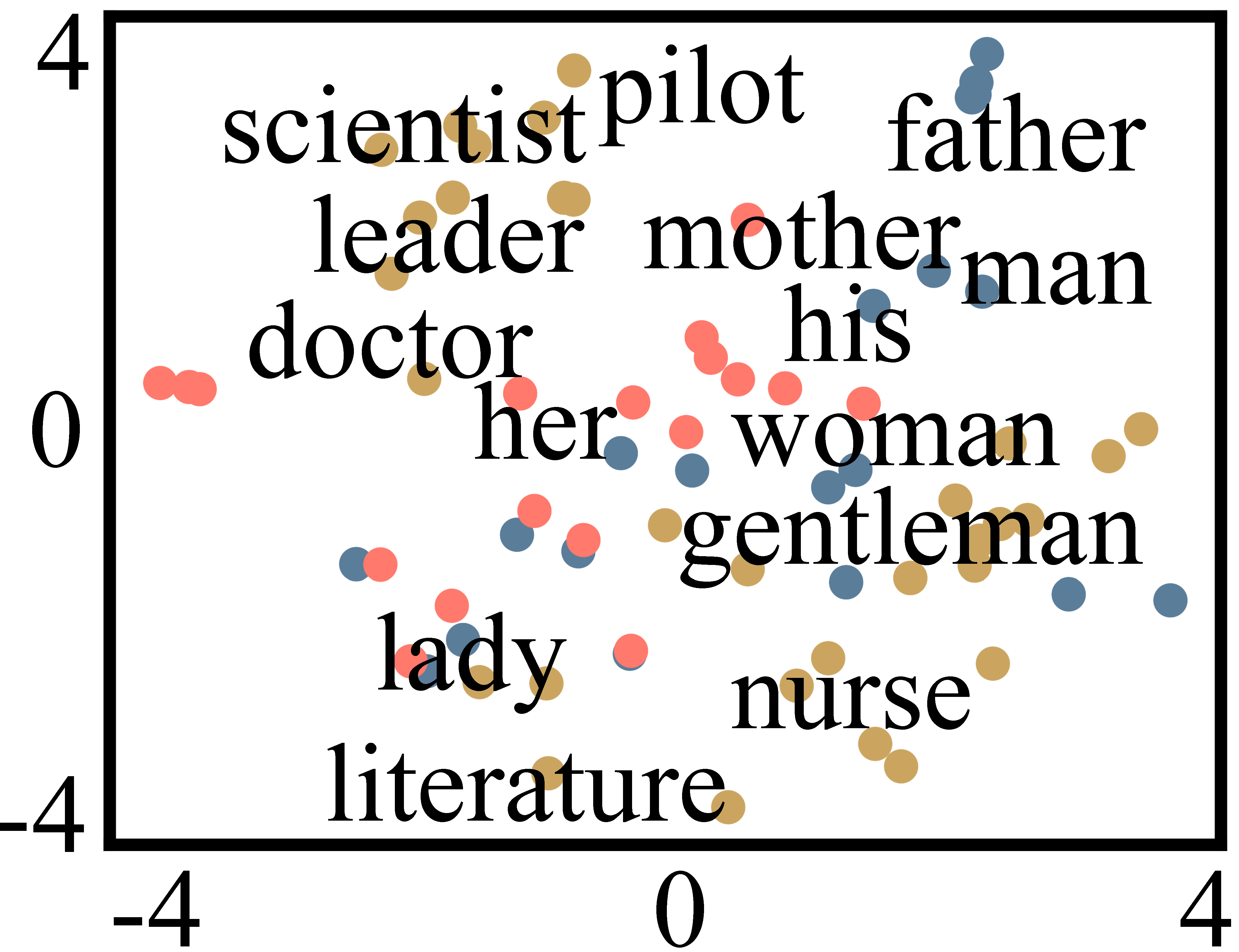}
}
\subfloat[Ours.]{
\label{fig:d}
    \setlength{\fboxrule}{1pt}
    \setlength{\fboxsep}{0pt}
	\includegraphics[width=0.13\textwidth]{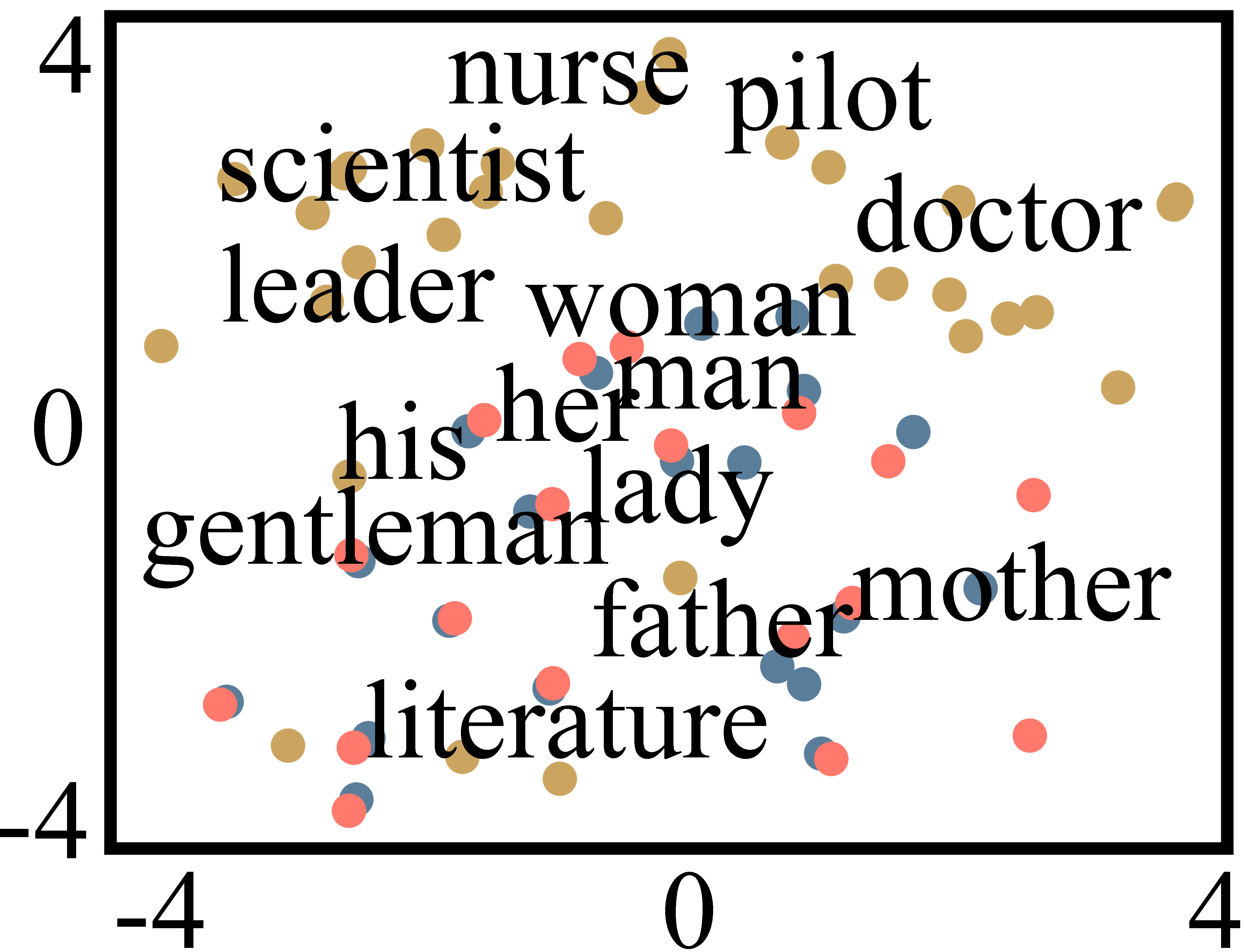}
}
\caption{$t$-SNE plots on BERT. Red: female, blue: male, and orange: neutral target words.}
\label{fig:tSNE}
\end{figure}

\noindent \textbf{Results on Language Understanding.}
Table~\ref{tab:GLUE-table} presents three GLUE results on debiased models. \M~slightly outperforms Auto-Debias on the CoLA and SST-2 tasks. This demonstrates that our $L_{Re}$ loss effectively addresses the common issue of reduced language understanding in most debiasing models~\cite{he2022mabel,sent-debias}. However, with the BERT backbone, \M~shows lower performance on the QNLI task compared to the task-aware SOTA model Causal-Debias. It's important to note that Causal-Debias integrates debiasing with downstream task fine-tuning, while our method is task-agnostic, which may make debiasing more challenging. Notably, our model does not reduce performance compared to the original BERT, and even improves Accuracy on SST-2.

The $t$-SNE visualization in Fig.~\ref{fig:tSNE} examines the debiasing effects and model expressiveness through words' correlation. Fig.~(\ref{fig:d}) shows \M~preserving relative distances and pulling attribute words together. While Auto-Debias has more dispersed distributions than the original BERT in Fig.~(\ref{fig:a}), Auto-Debias separates male and female words, suggesting that despite contextual similarities, opposing gender concepts remain distant in the hidden space, indicating persisting biases. 

\section{Conclusion}
In this paper, we offer a flexible, universally applicable solution \M~capable of debiasing lightweight PLMs by harnessing rich, social relevant pairwise sentences sourced from LLM, unlike existing methods reliant on crafted external corpora.
\M~roots in structural causal model (SCM) to reveal the limitations of direct utilization of generated sentences, such as alignment issues between LLM and PLMs, and negative knowledge transfer. 
We emoloy an improved causal graph to optimize the utilization of LLM-generated knowledge by filtering out sentences unaligned with PLMs, and only use aligned knowledge beneficial for positive transfer.
We rigorously conduct quality and toxicity tests for the generated sentences, which maintains their usability for the debiasing process.
Extensive evaluations show \M's efficacy in mitigating diverse biases across various PLMs, while also preserving model expressiveness when applied to a series of downstream tasks.

\section*{Acknowledgements}
This work was supported by Sichuan Provincial Science and Technology Program Projects (Grant No. 2023YFG0114 and 2024YFCK0003), Sichuan Provincial Science and Technology Program Projects (Grant No. 2024ZHCG0031), Science and Technology Program of Chengdu (Grant No. 2024YF0501231SN), Sichuan Province's Open Call for Proposals (Grant No. 2024YFCY0004), and Sichuan Province Central Leading Local Science and Technology Development Special Project (Grant No. 2024ZYD0265).
\newpage
\bibliographystyle{IEEEtran}
\bibliography{main}

\begin{thebibliography}{10}
\providecommand{\url}[1]{#1}
\csname url@samestyle\endcsname
\providecommand{\newblock}{\relax}
\providecommand{\bibinfo}[2]{#2}
\providecommand{\BIBentrySTDinterwordspacing}{\spaceskip=0pt\relax}
\providecommand{\BIBentryALTinterwordstretchfactor}{4}
\providecommand{\BIBentryALTinterwordspacing}{\spaceskip=\fontdimen2\font plus
\BIBentryALTinterwordstretchfactor\fontdimen3\font minus \fontdimen4\font\relax}
\providecommand{\BIBforeignlanguage}[2]{{%
\expandafter\ifx\csname l@#1\endcsname\relax
\typeout{** WARNING: IEEEtran.bst: No hyphenation pattern has been}%
\typeout{** loaded for the language `#1'. Using the pattern for}%
\typeout{** the default language instead.}%
\else
\language=\csname l@#1\endcsname
\fi
#2}}
\providecommand{\BIBdecl}{\relax}
\BIBdecl

\bibitem{bert}
J.~Devlin, M.-W. Chang, K.~Lee, and K.~Toutanova, ``Bert: Pre-training of deep bidirectional transformers for language understanding,'' in \emph{{NAACL}}, 2019, pp. 4171--4186.

\bibitem{roberta}
Y.~Liu, M.~Ott, N.~Goyal, J.~Du, M.~Joshi, D.~Chen, O.~Levy, M.~Lewis, L.~Zettlemoyer, and V.~Stoyanov, ``Roberta: A robustly optimized bert pretraining approach,'' \emph{arXiv:1907.11692}, 2019.

\bibitem{sun2019bias}
Y.~Sun, N.~Wang, X.-L. Shen, and X.~Zhang, ``Bias effects, synergistic effects, and information contingency effects: Developing and testing an extended information adoption model in social q\&a,'' \emph{Journal of the Association for Information Science and Technology}, vol.~70, no.~12, pp. 1368--1382, 2019.

\bibitem{elahi2021investigating}
M.~Elahi, D.~K. Kholgh, M.~S. Kiarostami, S.~Saghari, S.~P. Rad, and M.~Tkal{\v{c}}i{\v{c}}, ``Investigating the impact of recommender systems on user-based and item-based popularity bias,'' \emph{Information Processing \& Management}, vol.~58, no.~5, p. 102655, 2021.

\bibitem{shen2023towards}
T.~Shen, J.~Li, M.~R. Bouadjenek, Z.~Mai, and S.~Sanner, ``Towards understanding and mitigating unintended biases in language model-driven conversational recommendation,'' \emph{Information Processing \& Management}, vol.~60, no.~1, p. 103139, 2023.

\bibitem{meng2022generating}
Y.~Meng, J.~Huang, Y.~Zhang, and J.~Han, ``Generating training data with language models: Towards zero-shot language understanding,'' \emph{Advances in Neural Information Processing Systems}, vol.~35, pp. 462--477, 2022.

\bibitem{bhardwaj2021investigating}
R.~Bhardwaj, N.~Majumder, and S.~Poria, ``Investigating gender bias in bert,'' \emph{Cognitive Computation}, vol.~13, no.~4, pp. 1008--1018, 2021.

\bibitem{lu2022rationale}
J.~Lu, L.~Yang, B.~Mac~Namee, and Y.~Zhang, ``A rationale-centric framework for human-in-the-loop machine learning,'' \emph{arXiv preprint arXiv:2203.12918}, 2022.

\bibitem{veselovsky2305generating}
V.~Veselovsky, M.~Ribeiro, A.~Arora, M.~Josifoski, A.~Anderson, and R.~West, ``Generating faithful synthetic data with large language models: A case study in computational social science. arxiv 2023,'' \emph{arXiv preprint arXiv:2305.15041}.

\bibitem{WEAT}
A.~Caliskan, J.~J. Bryson, and A.~Narayanan, ``Semantics derived automatically from language corpora contain human-like biases,'' \emph{Science}, vol. 356, no. 6334, pp. 183--186, 2017.

\bibitem{kolisko2023exploring}
S.~Kolisko and C.~J. Anderson, ``Exploring social biases of large language models in a college artificial intelligence course,'' in \emph{Proceedings of the AAAI Conference on Artificial Intelligence}, vol.~37, no.~13, 2023, pp. 15\,825--15\,833.

\bibitem{mok2023people}
L.~Mok, S.~Nanda, and A.~Anderson, ``People perceive algorithmic assessments as less fair and trustworthy than identical human assessments,'' \emph{Proceedings of the ACM on Human-Computer Interaction}, vol.~7, no. CSCW2, pp. 1--26, 2023.

\bibitem{jatoba2019evolution}
M.~Jatob{\'a}, J.~Santos, I.~Gutierriz, D.~Moscon, P.~O. Fernandes, and J.~P. Teixeira, ``Evolution of artificial intelligence research in human resources,'' \emph{Procedia Computer Science}, vol. 164, pp. 137--142, 2019.

\bibitem{yu2023mixup}
L.~Yu, Y.~Mao, J.~Wu, and F.~Zhou, ``Mixup-based unified framework to overcome gender bias resurgence,'' in \emph{Proceedings of the 46th International ACM SIGIR Conference on Research and Development in Information Retrieval}, 2023, pp. 1755--1759.

\bibitem{mao2023debiasing}
Y.~Mao, L.~Yu, Y.~Yang, F.~Zhou, and T.~Zhong, ``Debiasing intrinsic bias and application bias jointly via invariant risk minimization (student abstract),'' in \emph{Proceedings of the AAAI Conference on Artificial Intelligence}, vol.~37, no.~13, 2023, pp. 16\,280--16\,281.

\bibitem{yu2024biases}
L.~Yu, L.~Guo, P.~Kuang, and F.~Zhou, ``Biases mitigation and expressiveness preservation in language models: A comprehensive pipeline (student abstract),'' in \emph{Proceedings of the AAAI Conference on Artificial Intelligence}, vol.~38, no.~21, 2024, pp. 23\,701--23\,702.

\bibitem{sent-debias}
P.~P. Liang, I.~M. Li, E.~Zheng, Y.~C. Lim, R.~Salakhutdinov, and L.-P. Morency, ``Towards debiasing sentence representations,'' in \emph{{ACL}}, 2020, pp. 5502--5515.

\bibitem{fairfil}
P.~Cheng, W.~Hao, S.~Yuan, S.~Si, and L.~Carin, ``Fairfil: Contrastive neural debiasing method for pretrained text encoders,'' in \emph{{ICLR}}, 2021, p.~1.

\bibitem{yang2023adept}
K.~Yang, C.~Yu, Y.~R. Fung, M.~Li, and H.~Ji, ``Adept: A debiasing prompt framework,'' in \emph{Proceedings of the AAAI Conference on Artificial Intelligence}, vol.~37, 2023, pp. 10\,780--10\,788.

\bibitem{auto-debias}
Y.~Guo, Y.~Yang, and A.~Abbasi, ``Auto-debias: Debiasing masked language models with automated biased prompts,'' in \emph{{ACL}}, 2022, pp. 1012--1023.

\bibitem{context-debias}
M.~Kaneko and D.~Bollegala, ``Debiasing pre-trained contextualised embeddings,'' in \emph{{EACL}}, 2021, p.~1.

\bibitem{he2022mabel}
J.~He, M.~Xia, C.~Fellbaum, and D.~Chen, ``Mabel: Attenuating gender bias using textual entailment data,'' \emph{arXiv:2210.14975}, 2022.

\bibitem{causal-debias}
F.~Zhou, Y.~Mao, L.~Yu, Y.~Yang, and T.~Zhong, ``Causal-debias: Unifying debiasing in pretrained language models and fine-tuning via causal invariant learning,'' in \emph{Proceedings of the 61st Annual Meeting of the Association for Computational Linguistics (Volume 1: Long Papers)}, 2023, pp. 4227--4241.

\bibitem{gender-tuning}
S.~Ghanbarzadeh, Y.~Huang, H.~Palangi, R.~C. Moreno, and H.~Khanpour, ``Gender-tuning: Empowering fine-tuning for debiasing pre-trained language models,'' \emph{arXiv preprint arXiv:2307.10522}, 2023.

\bibitem{zheng2023preserving}
J.~Zheng, Q.~Ma, S.~Qiu, Y.~Wu, P.~Ma, J.~Liu, H.~Feng, X.~Shang, and H.~Chen, ``Preserving commonsense knowledge from pre-trained language models via causal inference,'' \emph{arXiv preprint arXiv:2306.10790}, 2023.

\bibitem{chiang2023vicuna}
W.-L. Chiang, Z.~Li, Z.~Lin, Y.~Sheng, Z.~Wu, H.~Zhang, L.~Zheng, S.~Zhuang, Y.~Zhuang, J.~E. Gonzalez \emph{et~al.}, ``Vicuna: An open-source chatbot impressing gpt-4 with 90\%* chatgpt quality,'' \emph{See https://vicuna. lmsys. org (accessed 14 April 2023)}, 2023.

\bibitem{touvron2023llama}
H.~Touvron, L.~Martin, K.~Stone, P.~Albert, A.~Almahairi, Y.~Babaei, N.~Bashlykov, S.~Batra, P.~Bhargava, S.~Bhosale \emph{et~al.}, ``Llama 2: Open foundation and fine-tuned chat models,'' \emph{arXiv preprint arXiv:2307.09288}, 2023.

\bibitem{liang2024toa}
M.~Liang, Y.~Wu \emph{et~al.}, ``Toa: Task-oriented active vqa,'' \emph{Advances in Neural Information Processing Systems}, vol.~36, 2024.

\bibitem{li2024enhanced}
Y.~Li, R.~Zhang, J.~Liu, and G.~Liu, ``An enhanced prompt-based llm reasoning scheme via knowledge graph-integrated collaboration,'' \emph{arXiv preprint arXiv:2402.04978}, 2024.

\bibitem{bian2023chatgpt}
N.~Bian, X.~Han, L.~Sun, H.~Lin, Y.~Lu, and B.~He, ``Chatgpt is a knowledgeable but inexperienced solver: An investigation of commonsense problem in large language models,'' \emph{arXiv preprint arXiv:2303.16421}, 2023.

\bibitem{albert}
Z.~Lan, M.~Chen, S.~Goodman, K.~Gimpel, P.~Sharma, and R.~Soricut, ``Albert: A lite bert for self-supervised learning of language representations,'' in \emph{{ICLR}}, 2020, p.~1.

\bibitem{manzini2019black}
T.~Manzini, L.~Y. Chong, A.~W. Black, and Y.~Tsvetkov, ``Black is to criminal as caucasian is to police: Detecting and removing multiclass bias in word embeddings,'' in \emph{{NAACL}}, 2019, pp. 615--621.

\bibitem{SEAT}
C.~May, A.~Wang, S.~Bordia, S.~Bowman, and R.~Rudinger, ``On measuring social biases in sentence encoders,'' in \emph{{NAACL}}, 2019, pp. 622--628.

\bibitem{nadeem2020stereoset}
M.~Nadeem, A.~Bethke, and S.~Reddy, ``Stereoset: Measuring stereotypical bias in pretrained language models,'' \emph{arXiv preprint arXiv:2004.09456}, 2020.

\bibitem{crowspair}
N.~Nangia, C.~Vania, R.~Bhalerao, and S.~Bowman, ``Crows-pairs: A challenge dataset for measuring social biases in masked language models,'' in \emph{{EMNLP}}, 2020, pp. 1953--1967.

\end{thebibliography}
\end{document}